\documentclass[
]{ceurart}

\usepackage{amsmath}
\usepackage{amssymb}
\usepackage{caption} 
\usepackage{amsthm}
\usepackage{float}
\usepackage{setspace}
\usepackage{algorithm}
\usepackage{xcolor}
\usepackage[noend]{algpseudocode}
\newtheorem{theorem}{Theorem}
\newcommand{\qi}[1]{\textcolor{blue}{(#1)}}

\begin{document}

\copyrightyear{2021}
\copyrightclause{Copyright for this paper by its authors.\\
  Use permitted under Creative Commons License Attribution 4.0
  International (CC BY 4.0).}

\conference{In A. Martin, K. Hinkelmann, H.-G. Fill, A. Gerber, D. Lenat, R. Stolle, F. van Harmelen (Eds.), 
Proceedings of the AAAI 2021 Spring Symposium on Combining Machine Learning and Knowledge Engineering (AAAI-MAKE 2021) - 
Stanford University, Palo Alto, California, USA, March 22-24, 2021.}

\title{Knowledge Infused Policy Gradients for Adaptive Pandemic Control}

 \author{Kaushik Roy}
 \author{Qi Zhang}
 \author{Manas Gaur}
 \author{Amit Sheth}
 \address{Artificial Intelligence Institute, University of South Carolina}

\begin{abstract}
  COVID-19 has impacted nations differently based on their policy implementations. The effective policy requires taking into account public information and adaptability to new knowledge. Epidemiological models built to understand COVID-19 seldom provide the policymaker with the capability for adaptive pandemic control (APC). Among the core challenges to be overcome include (a) inability to handle a high degree of non-homogeneity in different contributing features across the pandemic timeline, (b) lack of an approach that enables adaptive incorporation of public health expert knowledge, and (c) transparent models that enable understanding of the decision-making process in suggesting policy. In this work, we take the early steps to address these challenges using Knowledge Infused Policy Gradient (KIPG) methods. Prior work on knowledge infusion does not handle soft and hard imposition of varying forms of knowledge in disease information and guidelines to necessarily comply with. Furthermore, the models do not attend to non-homogeneity in feature counts, manifesting as partial observability in informing the policy. Additionally, interpretable structures are extracted post-learning instead of learning an interpretable model required for APC.  To this end, we introduce a mathematical framework for KIPG methods that can (a) induce relevant feature counts over multi-relational features of the world, (b) handle latent non-homogeneous counts as hidden variables that are linear combinations of kernelized aggregates over the features, and (b) infuse knowledge as functional constraints in a principled manner. The study establishes a theory for imposing hard and soft constraints and simulates it through experiments. In comparison with knowledge-intensive baselines, we show quick sample efficient adaptation to new knowledge and interpretability in the learned policy, especially in a pandemic context.
\end{abstract}


\maketitle

\section{Introduction}
Reinforcement learning (RL) is one of the main techniques to solve sequential decision making problems.
When combined with deep neural networks, RL has achieved impressive performance in many applications, including robotics \cite{kober2013reinforcement}, game playing \cite{silver2016mastering}, recommender systems \cite{li2010contextual}, etc.
As RL fundamentally solves decision making problems via trial and error, a major drawback of RL is the huge amount of interactions required to learn good decision policies, which can lead to prohibitive cost and slow convergence.

The inefficiency of RL has motivated studies on incorporating expert domain knowledge when solving decision making problems.
In this direction, prior work has largely focused on the setting of imitation learning (IL), in which knowledge is represented through expert demonstrations, i.e., the expert demonstrates the desired behavior (rather than specifying the reward signal) in various scenarios stemming from a decision making problem.
Such demonstrations can be used to either directly learn a classifier that mimics the expert's behavior, known as behavior cloning \cite{ross2011reduction}, or infers the reward function that rationalizes the expert's behavior and is then optimized through RL, known as inverse RL \cite{abbeel2004apprenticeship}.
The performance of IL relies on the quality of expert demonstrations.
Expert demonstrations are often not exhaustive to provide supervision for all kinds of scenarios that might be countered.
Moreover, sometimes only a suboptimal expert is available due to human's bounded rationality. Therefore policies learned with IL are inferior to the policies learned with RL that uses the problem's original reward.
These limitations of IL has further motivated recent work that attempts to combine RL and IL.
These approaches leverage both the original reward and expert demonstrations to learn better-than-expert policies faster than RL-only approaches.
For example, AlphaGo is pretrained with human expert moves and then refined via RL \cite{silver2016mastering}.

This paper focuses on combining RL and expert knowledge to solve sequential decision making problems, which we term as knowledge infused RL.
Different from the aforementioned prior works, we aim to deal with the following important, yet understudied challenges in knowledge infused RL (KIRL).
1) Partial observability and non-stationarity. Many real-world domains modeled as sequential decision making problems are partially observable. Moreover, there are often exogenous factors that are un-modeled in the input features, making the underlying decision making problem non-stationary. Most existing works in KIRL (e.g., combining RL and IL) ignore this and apply algorithms developed with the assumption of full observability and stationarity, while we aim to explicitly consider partial observability and non-stationarity in our KIRL method. 
2) Structured knowledge representation. The IL framework assumes that knowledge is represented by expert demonstrations of low-level behavior, yet low-level demonstrations are difficult to obtain in many domains. Instead, human knowledge is often represented in a structured, high-level manner. As an example, consider the guidelines of a public health agency for a pandemic. We argue that leveraging such structured knowledge requires structured input representation for the RL algorithm.  
3) Interpretability. We aim to develop approaches such that both the process of leveraging expert knowledge and the resulting learned policy are interpretable. The interpretability for KIRL is required in high-stakes domains such as public health and yet largely ignored in prior works.

\paragraph{Adaptive Pandemic Control}
This paper focuses on the pandemic control setting that manifests all of the three aforementioned challenges.
Consider a scenario where a notional city is in a pandemic, with the following characteristics: people living in homes, working at offices, and shopping at places, all connected in a geographical map.
The city's government aims to optimize its pandemic control policy with KIRL to strike a balance between public health and economic resiliency. 
The true number of infected people is only partially observable to the government, as exhaustive testing is often not possible.
In general, the course of the pandemic is a non-stationary process due to exogenous factors, such as people flowing into a certain area for some short-term local event, an abrupt decrease in the mortality rate of the disease due to new medications, etc. The partial observability and non-stationarity result in a non-homogeneous counting process of the observed number of infected individuals.
The knowledge of experts consulted is often formatted into high-level guidelines; for example, shopping area A should be locked down before shopping area B since locking down B will result in more severe economic consequences.
In order to incorporate such guidelines, the input to the KIRL algorithm needs to be represented in an interpretable manner. Further, the resulting pandemic control policy learned by KIRL needs to be also interpretable.

\paragraph{Related Work}
For APC, during COVID-19, we have identified specific challenges that include the agent handling relational features, non-homogeneity in the feature counts, and learning non black-box interpretable structures through knowledge infusion \cite{gaur2020knowledge}. We believe that the inability to handle these issues by previous approaches can pose bottlenecks in agent models for assisting policymakers. Our study aims to investigate our formulation in handling these specific challenges through knowledge infusion in functional space since we specify knowledge as functional constraints. There is a rich body of work on RL concerning relational feature-based functional spaces \cite{kersting2008non, das2020fitted}. However, they do not use knowledge infusion, handle count features, or deal with partial observability in the state. Poisson dependency networks have been proposed to handle multivariate count data but do not consider non-homogeneity in the counts \cite{hadiji2015poisson}. Odom et al., present a way to incorporate knowledge constraints in relational function spaces that can be used in conjunction with the work of Kersting et al., and Hadiji et al., to achieve knowledge infusion in Policy Gradients \cite{odom2015knowledge, kersting2008non, hadiji2015poisson} for pandemic control. Our work most closely resembles this, and we, therefore, employ it as our evaluation baseline. We make key and necessary modifications to the agent's approximation architecture in moving from trees to linear basis, using kernel aggregates to handle non-homogeneity, partial observability and most importantly, development of a mathematical framework for hard and soft imposition of knowledge as functional constraints that are applicable in a wide range of scenarios in APC. Also, we prove that the baseline is an instance of our framework. Other approaches for knowledge infusion in functional spaces include the use of cost sensitivity constraints in imbalanced data, and monotonicity constraints which are not directly applicable to our setting \cite{yang2014learning, kokel2020unified}. The use of monotonicity constraints in preference-based knowledge infused RL can be an interesting extension to our work.

\begin{figure}
    \centering
    \includegraphics[width=\textwidth]{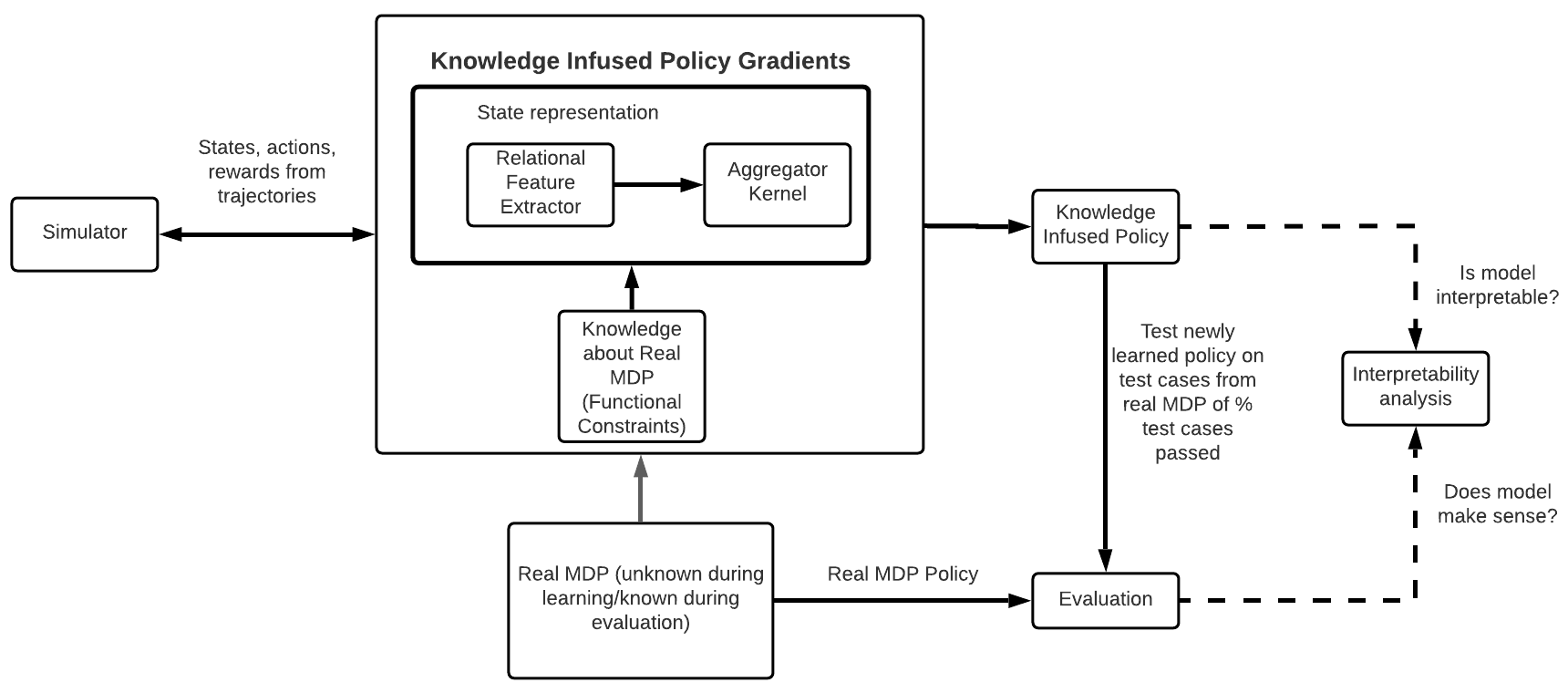}
    \caption{The knowledge infusion and analysis pipeline that begins with the agent interacting with the simulator to learn the policy, constructing a state representation and utilizing knowledge through KIPG to learn a knowledge infused agent policy for APC scenarios. This policy is then evaluated for its efficacy (\% of test cases passed) and interpretability (readable and meaningful).}
    \label{fig:my_labelPipeline}
\end{figure}

\paragraph{Contribution}
Our contribution is two folds:
First, we create an agent-based pandemic simulator that models the interactions between individuals that move across specific locations within a community, such as homes, offices, shops, hospitals, etc.
The spread of the disease is simulated using the typical SIR model.
Interventions like locking down a specific location and increasing testing are the control measures modeled in the simulator.
The pandemic simulator manifests all of the three motivating challenges.
Second, we develop a novel KIRL algorithm, Knowledge Infused Policy Gradient, that addresses the challenges.
To incorporate structured knowledge format and support interpretability, the policy is derived from learned relational features using an interpretable 2-layer neural network. 
An example of such a relational feature is - \textit{There exists a residential neighborhood, a person living in a home, and shop in the same route with many people shopping at this place, where a potential intervention is locking down such shops.}
The partial observability is addressed by aggregating the learned relational features over time.
Further, expert knowledge is infused into the policy gradient-based optimization in an online manner so that the knowledge can be adjusted whenever necessary to adapt to the non-stationary course of the pandemic.    
Figure \ref{fig:my_labelPipeline} shows the pipeline of our KIRL algorithm, which will be described in detail in Sections \ref{sec:Preliminaries} and \ref{sec:M}.


\section{Preliminaries}
\label{sec:Preliminaries}
\subsection{Policy Gradients in Functional Space}

In the standard RL framework, a Markov Decision Process (MDP) is defined by a set of states $s\in S$, actions $a\in A$, state transition probabilities on taking actions $\delta(s,a,s^{'}): S \times A \times S \rightarrow [0,1]$, and a reward model $r(s,a):S \times A\rightarrow \mathbf{R}$. A common way to specify the policy $\pi(s,a)$, to execute in state $s$, is using a Boltzmann distribution:  $\pi(s,a) = \frac{e^{\psi(s,a)}}{\sum_{a'}e^{\psi(s,a^{'})}}$, where $\psi(s,a) = \theta^{\mathbf{T}}\Phi(s,a)$, where $\Phi(s,a)$
represents features about $(s,a)$, $\theta$ are the parameters and $a^{'} \in A$. Policy gradients seek to learn the parameters $\theta$, that optimize the value of a policy:
\begin{equation}
  \frac{\partial\nu_{\pi}}{\partial\theta} = \sum_{s}d^{\pi}(s)\sum_{a}\frac{\partial\pi(s,a)}{\partial\theta}.Q^{\pi}(s,a) 
  \label{eq:1}
\end{equation}
Here, $d^\pi$ is the distribution from which the states are drawn and $Q^\pi$ is the Q-value function corresponding to policy $\pi$.
The $Q^{\pi}$(s,a) function is estimated using state-action pair trajectories from a simulator, by Monte Carlo methods or function approximation.
Kersting et al., show that $\theta$ parameterization of the policy is difficult to achieve owing to feature selection in continuous and relational environments, in which there are infinitely many possibilities \cite{kersting2008non}. Thus we employ gradient ascent in functional space to learn the function $\psi(s,a)$ directly, that rely on learned relational features (see Section \ref{sec:RFE and Aggregator Kernel}). The learning of these features overcomes the problem of pre-defining count features, and thereby providing finer grained control. We start with an initial function $\psi_{0}(s,a)$ and add $k = 1 - K$ functions, $\delta_{k}(s,a)$ to fit the gradients: 
$\partial\nu_{\pi}/\partial\psi_{k-1}(s,a)$
where $\psi_{k}(s,a) = \psi_{0}(s,a) + \sum_{j=1}^{k-1}\delta_{j}(s,a)$.

In the parametric setting we use the gradient specified in Equation \ref{eq:1}. Here we make the change from $\theta$ to the function $\psi(s,a)$, we need the form for the gradient $\frac{\partial\pi(s,a)}{\partial\psi(s,a)}$, as this is the only component of the gradient dependent on $\psi(s,a)$, for each action $a \in A$. We instead compute the gradient as:
$
\frac{\partial\pi(s,a)}{\partial\psi(s,a)} \equiv \pi(s,a)\frac{\partial \log\pi(s,a)}{\partial\psi(s,a)}
$.
Using the Boltzmann distribution form for $\pi(s,a) = \frac{e^{\psi(s,a)}}{\sum_{a'}e^{\psi(s,a^{'})}}$, this gradient becomes: 
$\frac{\partial \log\pi(s,a)}{\partial\psi(s,a)} = I(s,a) - \pi(s,a)$. 
This has a very intuitive form, as it defines the gradient between, if $a$ was taken in $s$ given by the indicator function $I(s,a)$, and the probability of taking $a$ in $s$ according to our model given by $\pi(s,a)$.

\subsection{Gradient boosted 2-layer Neural Network Learning}
In this paper, the agent uses a Neuro-Symbolic approximator for the policy by learning a set of linear models. Each linear model $\delta_k(s,a)$ can be viewed as a basis function for $\psi(s,a)$ approximation. To prevent the correlation between them, the sample state action pairs will be sub-sampled each time. The resulting policy will be a linear combination of these basis functions. We move away from tree-based models, for the following reasons:
1) The basis functions can handle continuous, discrete, and relational inputs without extensive pre-processing or binning.
2) The final linear combination $\psi(s,a)$ can be laid out as a 2-layer Neural Network(NN), for which the weights in the 2nd layer are unity. The network weights can be refined through backpropagation. Beubeck et al., show that a 2-Layer NN with RELU activations are able to approximate arbitrary functions with high precision and $O(\frac{N}{d})$ neurons in the hidden layer, where $d$ is the dimension of the input and $N$ is the number of data samples \cite{bubeck2020network}. Since the dimensionality $d$ in our setting is high and $N$, i.e. size of the data is typically low, we can use just a few linear basis functions (neurons) in the hidden layer to achieve good approximation.
3) Each basis function $\delta_k(s,a)$ has interpretable structure. Interpretability theory for NN structures has recently been well studied and we require that the agent policy be robust and interpretable for applications such as APC. Dombrowski et al., show that a target interpretation on the decision making by the NN $\mathcal{E}_{\pi(s,a)}$, given by the network weights (heat map), can be manipulated in terms of its features $\Phi(s,a)$ and yet still yield almost the same interpretation $\mathcal{E}_{\pi(s,a)}$ \cite{dombrowski2019explanations}. This is related to the curvature of the output manifold of the NN. They propose to alleviate this issue by replacing RELU activations with soft-plus non linearities with a small parameter $\beta$ as: $\frac{1}{\beta}\log(1+\beta\Phi(s,a))$. With this modification, the weights are more robust to perturbations or modifications in the input $(s,a)$.

\section{Methodology}
\label{sec:M}

\subsection{Relational Feature Extractor and Aggregator Kernel}
\label{sec:RFE and Aggregator Kernel}
We learn the relational features (clauses) over of the state $\Phi(s,a)$ using standard and well understood Inductive Logic Programming (ILP) methods \cite{muggleton1992inductive}. The inductive bias is provided in the form of Aleph modes which can be automatically learned from a schema of the world \cite{hayes2017user, srinivasan2001aleph}. This bias is included to constraint the search to not include features that do not make sense. For example, to decide on locking down a shop we would not like an irrelevant feature that does not actually contain the shop or anything related to it. Furthermore, we count the number of examples over the features that make up $\Phi(s,a)$ for example the feature: 
\textbf{same(State,Res,Shop)} $\land$ \textbf{pin(State,Person,Home)} $\land$ \textbf{hin(State, Home,Res)} outputs the number of persons, homes, shops, and residential areas that satisfy this feature description, where the feature denotes: \textit{There exists a residential neighborhood and shop in the same route and a person living in a home that is part of the residential neighborhood}. We make use of a minimal threshold of mutual information to satisfy along with a maximum clause length for picking the features $\Phi(s,a)$.

It is likely in APC, that $\pi(s,a)$ is a process that depends on counts $\lambda_{h(s,a)}$ over latent state features $h(s,a)$, such as number of persons ill i.e. $\lambda_{\Phi(s,a)} = f(\lambda_{h(s,a)})$. This creates partial observability, and thus the entire observed data trajectory history $\mathcal{H} = \{(s_{1},a_{1}),(s_{2},a_{2}) ..$ $(s_{T},a_{T})\}$ influences counts $\lambda_{\Phi(s,a)}$ and in turn the policy $\pi(s,a)$. Hadiji et al., present a way to handle multivariate count models, although not latent counts \cite{hadiji2015poisson}. However, their model does not take into account non-homogeneity in the counts that is also characteristic of APC scenarios. Hence, due to expected non-homogeneity, the count can be modeled as: \[\lambda_{\Phi(s,a)}^T = \mu_{T} + \sum_{t=1}^{T-1}w_{t}\mathcal{K}(\lambda_{\Phi(s,a)}^{t},\lambda_{\Phi(s,a)}^{T-1}),\] 
where $\mu_{T}$ models a base count (bias).
We aggregate count features that make up over the history $\mathcal{H}$ using this type of kernel $\mathcal{K}$ to handle partial observability and non-homogeneity in the counting process. The Kernel approach inspired by a Hawkes process model \footnote{\url{https://mathworld.wolfram.com/HawkesProcess.html}}, acts as a method to model a homogeneous count process in the local neighborhood around $T$, where the non-homogeneity arises due to a union of several such locally homogeneous processes. Additionally, it has been studied before that the limiting distribution of a binary outcome process over an infinite horizon is a counting process, for example binomial to Poisson \cite{simons1971convergence}. We can similarly use the same principle in reverse to model the policy as a binary outcome process over each discrete $(s,a)$, at time $T$. We thus model the policy $\pi(s,a)$ at time $T$ as a binary outcome problem, using relational count features from the trajectory history $\mathcal{H}$, up to time $T-1$, aggregated using kernel $\mathcal{K} = e^{-(x-y)^{2}}$

\subsection{Bayesian Knowledge Infusion}
Using a relational description, we can mathematically formalize knowledge as specifying constraints over the parameters $\theta$. More precisely, knowledge is specified as a set of $M$ functional constraints $\mathit{FC_{i}}$, where $i \in [1,M]$ for each action $a \in A$ as: $\land f_{i}(s) \Rightarrow (P(\theta) = p_{i}(\omega_{i},\theta))$, which says that functional constraints are applied in probability space to the parameters, if the conjunction of conditions, $\land f_{i}$ are satisfied in state $s$. If $P(\theta)$ is high, then action $a$ is preferred in $s$ when $\land f_{i}$ applies. In functional space, $\psi(s,a)$ is constrained in place of $\theta$. We will denote $\mathbf{D}$ to be the data containing $(s,a)$ pairs from trajectories.
 $\psi(s,a)$ is updated according to Bayes rule that defines the posterior as: 
 \[
 P(\psi(s,a)|\mathbf{D}) =   \frac{P(\mathbf{D}|\psi(s,a))P(\psi(s,a))}{\int_{\psi(s,a)}P(\mathbf{D}|\psi(s,a))P(\psi(s,a))}
 \]
 
Taking $\log$ on both sides we get 
$
\log P(\psi(s,a)|\mathbf{D}) \propto \log P(\mathbf{D}|\psi(s,a))+ \log P(\psi(s,a))
$.
For our problem the distribution $P$ is the policy $\pi$, that we trying to learn. Therefore, $P(\mathbf{D}|\psi(s,a)) = \pi(s,a) = \frac{e^{\psi(s,a)}}{\sum_{a'}e^{\psi(s,a^{'})}}$, and $\log P(\psi(s,a))$ are the $\omega_{i}$'s corresponding to each $FC_{i}$, of which there are $M$. We assume independence among the $FC_{i}$'s. Instead of using the data likelihood $P(\mathbf{D}|\psi(s,a))$ as the functional form of policy $\pi$, we now use the Bayesian posterior, $P(\psi(s,a)|\mathbf{D})$. This gives us a new form for policy:
$
\log\pi(s,a) \propto \log(\frac{e^{\psi(s,a)}}{\sum_{a'}e^{\psi(s,a^{'})}}) + \log P(\psi(s,a))
$.
Using the Laplace distribution form i.e. $p_{i}(\psi(s,a), \omega_{i}) = \frac{e^{\frac{-|\psi(s,a)-\omega_{i}|}{b}}}{2b}$  and setting $b=1$, we now derive the new knowledge infused functional gradient $\frac{\partial \log\pi(s,a)}{\partial\psi(s,a)}$, when $\mathit{FC_{i}}$ applies in $s$ when learning model for action $a$, as follows:
\begin{equation} \label{eq:2}
\begin{split}
\frac{\partial \log\pi(s,a)}{\partial\psi(s,a)} = \frac{\partial\log(\frac{e^{\psi(s,a)}}{\sum_{a'}e^{\psi(s,a^{'})}}) + \log P(\psi(s,a))}{\partial\psi(s,a)}
 = I(s,a)-P(\mathbf{D}|\psi(s,a) -\mathbf{sign}(\psi(s,a) -\omega_{i}) 
\end{split}
\end{equation}
where, $-\mathbf{sign}(\psi(s,a) -\omega_{i})$ is $1$ when $FC_{i}$ really prefers $a$ in $s$, with $\omega_{i}$ being a large positive number, and is $-1$ when $FC_{i}$ does not prefer $a$ in $s$, with $\omega_{i}$ being a large negative number. With multiple $FC_{i}$, we can write the knowledge infused functional gradient as:
\begin{align*}
(I(s,a)-P(\mathbf{D}|\psi(s,a))) + \sum_{i}\alpha_{i}(-\mathbf{sign}(\psi(s,a) -\omega_{i}))
\end{align*}
where $\alpha_{i}$ can be thought of as a weight on how important we consider $FC_{i}$. It is worth noting that if we set all $\alpha_{i} = \alpha$ in Equation \ref{eq:2}, we recover the formulation in Odom et al.'s work \cite{odom2015knowledge}.
We formally state this in Theorem \ref{thm:Odom}, where the proof is omitted due to the space limit. 
More generally, $p_{i}(\psi(s,a),\omega_{i})$ can assume functional forms other than Laplace distributions as well, depending on domain requirements. 
\begin{theorem}
\label{thm:Odom}
With $\omega_{i}$ for each Functional Constraint set to $\pm K$ where $K$ is the number of learned basis functions during Functional Gradient Ascent, and all $\alpha_{i} = \alpha$, 
\begin{equation}\label{eq:3}
    \begin{aligned}
(I(s,a)-P(\mathbf{D}|\psi(s,a))) + \sum_{i}\alpha_{i}(-\mathbf{sign} (\psi(s,a) -\omega_{i})) =  (I(s,a)-P(\mathbf{D}|\psi(s,a))) + \alpha (n_t - n_f)
\end{aligned}
\end{equation}
$n_t$ is the number of Functional Constraints that agree with the action taken and $n_f$ is the number that does not.
\end{theorem}

\subsection{Conditional Functional Gradients for Knowledge Infusion}
The Bayesian formulation imposes constraints in a soft way such that the agent gradually incorporates the knowledge. A second way to incorporate knowledge as functional constraints that enforces hard imposition, is to use the conditional functional gradient ascent method \cite{wang2015functional}. In this method, after $\psi_{k}(s,a)$, after $k$ stages of boosting, is approximated as $\psi_{0}(s,a)+\sum_{j=1}^{k-1}\delta_{j}$, a constrained $\psi_{k}(s,a)$ is obtained by solving the linear program (LP) \[\psi_{k}^{*}(s,a) =\mathbf{argmin}_{\psi_{FC_{i}}(s,a)} \psi^{\mathbf{T}}(s,a)(\pi(s,a)\frac{\partial \log\pi(s,a)}{\partial\psi(s,a)}Q(s,a))\], where $\psi_{FC_{i}}(s,a)$ are the $\psi(s,a)$ functions that are constrained to adhere to $FC_{i}$, and $\pi(s,a) = \frac{e^{\psi(s,a)}}{\sum_{a'}e^{\psi(s,a^{'})}}$. $\psi_{k}(s,a)$ is then recomputed as $(1-\gamma_{k})\psi_{k}(s,a) + \gamma_{k}\psi_{k}^{*}(s,a)$, where $\gamma_{k} \in [0,1]$. $\gamma_{k}$ is a hyper-parameter that is empirically set and decayed as learning progresses. The form of the $FC_{i}$, is the same as in the Bayesian formulation. Thus these constraints can be infused into the optimization through conditional functional gradients to handle $\mathbf{Source, Cost~Sensitivity}$ and $\mathbf{Hybrid}$ constraints as well. Any off the shelf LP solver can be used during optimization.The LP formulation for $FC_{i}$ is detailed below:
\begin{equation}
\begin{split}
   \min~\sum_{(s,a)\in\mathbf{D}}(\psi_{k}(s,a))(\pi(s,a)\frac{\partial \log\pi(s,a)}{\partial\psi(s,a)}Q(s,a)) \quad \mathrm{s.t.}~~
   \psi_{k}(s,a) = \omega_{i},~\mathrm{if}~\land(f_{i}(s)=\mathbf{True})
\end{split}
\end{equation}
The LP can also be solved using gradient descent on the Lagrangian constructed as:
if $\land(f_{i}(s)=\mathbf{True})$
\begin{equation*}
    \begin{aligned}
    \mathcal{L}(\psi(s,a)) = \sum_{(s,a)\in\mathbf{D}}\psi_{k}(s,a)(\pi(s,a)\frac{\partial \log\pi(s,a)}{\partial\psi(s,a)} Q(s,a))
    - \alpha_{i} (\psi_{k}(s,a)-\omega_{i})
\end{aligned}
\end{equation*}
\subsection{Combination of Hard and Soft Constraints}
In APC, assistive agents are required to comply with general guidelines (hard constraints) while also benefiting from adapting to knoweldge in a gradual manner (soft constraints).  We can combine the Bayesian formulation with Conditional Functional Gradients to achieve this type of agent. Thus, first knowledge is specified for soft infusion using $FC_{soft} = \land f_{soft}(s) \Rightarrow (P(\psi(s,a)) = p_{soft}(\omega_{bias},\psi(s,a)))$, following which $\frac{\partial \log\pi(s,a)}{\partial\psi(s,a)}$ is computed as $(I(s,a)-P(\mathbf{D}|\psi(s,a))) + -\mathbf{sign}(\psi(s,a) -\omega_{soft})$.  Next, hard constraints that the agent has to comply with can be specified for hard infusion using $FC_{hard}$ $\land f_{hard}(s) \Rightarrow (P(\psi(s,a)) = p_{hard}$ $(\omega_{hard},\psi(s,a)))$, which can be optimized using an LP solver or by solving the Lagrangian:
if $\land(f_{hard}(s)=\mathbf{True})$
\begin{equation*}
    \begin{aligned}
    \mathcal{L}(\psi(s,a)) = \sum_{(s,a)\in\mathbf{D}}(\psi_{k}(s,a))(\pi(s,a)\frac{\partial \log\pi(s,a)}{\partial\psi(s,a)} Q(s,a))
    - \alpha_{hard} (\psi_{k}(s,a)-\omega_{hard})
\end{aligned}
\end{equation*}
In this way, knowledge infusion is carried for the agent to effectively assist policy makers with both incoming knowledge about the developing situation and compliance with general guidelines.
\section{Experiments and Evaluation}
We design a simulator that interacts with the agent. It simulates the pandemic spread in a small city. The simulator includes information about persons, households, and facilities in the city namely - residential areas, hospitals, shops, and workplaces. Actions are lock/unlock parts of the city as well as increase testing by 10\%. 
The reward model is to reduce human fatalities. 
We now compare the knowledge infusion methods, with and without feature aggregation.
Table \ref{tab:relational-features} shows feature examples that stayed consistent after the ILP module across all tasks and hence were retained for conducting the experiment, where \textit{same} denotes that the establishments are along the same route, \textit{pin} denotes person in home, \textit{hin} denotes home in residential neighborhood, \textit{sopen} denotes an open shop, \textit{hostpitalized} and \textit{quarantined} denotes persons hospitalized or quarantined, \textit{ropen}, \textit{wopen} and \textit{hopen} denotes residential neighborhood, workplace and home being open. Here open means not placed under lockdown. Possible actions are to lockdown or unlock routes, homes, residential neighborhoods, shops, or workplaces and increase testing at these locations. "NilPolicy" is also included as part of the action space. All features are existentially quantified.\newline

\begin{table}
\centering
\begin{tabular}{|p{0.3cm}|p{7cm}|p{6cm}|}
 \hline
ID & Feature & Description of Clause \\ [0.5ex] 
 \hline\hline
 1 & same(State,Res,Shop)$\land$pin(State,Person,Home)$\land$ hin(State,Home,Res) &  There exists a residential neighborhood and shop in the same route and a person living in a home that is part of the residential neighborhood.\\ 
 \hline
 2 & same(State,Shop,Work)$\land$ pin(State,Person,Home)$\land$ hin(State,Home,Res)$\land$same(Shop,Res,Work) & There exists a shop, a workplace, and a residential neighborhood in the same route and person living in a home belonging to that residential neighborhood.\\ [1ex]
 \hline
 3 & sopen(State,Shop) & There exists a Shop that is open.\\ 
 \hline
 \hline
\end{tabular}
\caption{\label{tab:relational-features}Some relational features that are learned with their English descriptions. It can be seen that the features allow finer grained control at the level of individual shops, homes, residences, workplaces, and routes.}
\end{table}

\subsection{\textbf{ Comparison of Knowledge Infusion methods}}
The agent learns by interacting with the simulator to optimize the reward of minimizing infections. It is then subsequently required to adapt to new knowledge about the city map. This knowledge is provided in the form of prohibition of closing down of certain parts of the city specified as functional constraints with $\omega$ denoting the constraint strength/importance adjudged by the constraint specifier, for example, lockshop(State, Shop):-sopen(State,Shop), -1 and $\alpha = 1$ denoting the confidence of incorporation in terms of trust/validity for the bayesian infusion technique. For the combined setting, we also use lockshop(State,Shop):-ph(State,Person), $+1$ for infusion by Conditional Gradients. The aim is to combine the hard constraint of priority being given to locking down places of interaction such as shops, if many people are hospitalized as there are now fewer beds and the soft constraint of keeping open shops running in order to keep the economy functioning. Table \ref{tab: KI-Compare} shows how Knowledge Infusion using all methods, fares against Policy Gradient without Knowledge Infusion (KI) and against the baseline which use relational count features and combines Odom et al.,\cite{odom2015knowledge}'s knowledge infusion with Kersting et al.,\cite{kersting2008non}'s Policy Gradient approach. The baseline also uses linear basis instead of trees. 
We define a test-case passed as the number of times policy choice is equal to the \emph{real MDP} choice. Recall from \textbf{Figure} \ref{fig:my_labelPipeline} that the \emph{real MDP} is known during evaluation. The percentage of test cases passed is reported against number of simulator trajectories. Note that in the combined setting, the baseline approach which uses Odom's KI cancels out the effect of knowledge that is weighted contrastingly i.e. $\alpha$ and $-\alpha$. Also, as seen in Section \ref{sec:RFE and Aggregator Kernel}, aggregation is used to model Partial Observability in the state. We note that aggregation shows improved performance as without aggregation the \% test-cases passed are on average 10\% lower than with for 20,50 and 100 trajectories, across all comparison settings.
\begin{table}
\begin{center}
 \begin{tabular}{|c|c|c|c|c|c|c|} 
 \hline
 Trajectories & Bayes & CFG & Co & w/o & B & B-Co\\ [0.5ex] 
 \hline\hline
 20 & 0.8 & 0.85 & 0.85 & 0.4 & 0.79 & 0.5\\ 
 \hline
 50 & 0.9 & 0.95 & 0.85 & 0.6 & 0.9 & 0.65\\
 \hline
 100 & 0.94 & 0.95 & 0.85 & 0.7 & 0.93 & 0.7\\ [1ex] 
 \hline
\end{tabular}
\end{center}
\caption{\label{tab: KI-Compare}Comparison of Different types of knowledge infusion to test sample efficiency. Bayes: Bayesian, CFG: Conditional Functional Gradient, Co: Combined, w/o: Without, B: Baseline, B-Co: Baseline Combined}
\end{table}
It can be seen that the sample efficiency vastly improves with KI than without. More over, in other settings the baseline is similar in results because it is exactly derivable from the bayesian formulation as proved in Theorem \ref{thm:Odom}.
\subsection{Exogeneity of Multiple Events}
We define an \emph{Event} as - setting into movement, a certain population of people who don't follow the typical simulation dynamics. For example, as already mentioned, the simulator encodes that persons in a residential area shop at a "shop" that is in the neighborhood. But we select certain persons who may instead deviate from this routine. A concrete example of this might be those who deviate from their daily routine of going to work to instead gather at a location staging a rally. This \emph{Event} if not handled early, for example by locking down the location and testing everyone that attended, can cause unintended consequences in terms of both human fatalities and economic losses incurred. We demonstrate how knowledge infusion can be used to mitigate this effect. 
\begin{table}
\begin{center}
 \begin{tabular}{|c|c|c|c|} 
 \hline
Method & \textbf{Event 1} & \textbf{Event 2} & \textbf{Event 3} \\ [0.5ex] 
 \hline\hline
 KIPG-Bayesian with $\lambda=1$ & 2 & 2 & 2 \\ 
 \hline
 KIPG-Bayesian with $\lambda = 0.5$ & 3 & 3 & 3 \\
 \hline
 KIPG-Bayesian with $\lambda=2$ & 1 & 1 & 1 \\  
 \hline
KIPG-CFG & 2 & 2 & 2 \\ [1ex]
\hline
\end{tabular}
\end{center}
\caption{\label{tab: multi-event-KI}Comparison of different configurations of knowledge infusion when multiple events are injected into the simulation. The time taken by each method to reach $\geq 0.75$ \% test cases passed is recorded.}
\end{table}
We must take care here to analyse the \emph{Event} before acting too quickly in infusing knowledge. Since the effect of imposing policy in RL at any given time step propagates forward to all other time steps, in the case that the \emph{Event} is transient in nature (passes quickly), knowledge infusion may cause more harm than good.
\subsection{Interpretability Analysis}
We analyze the weights at the input layer of the neural network to understanding which parts of the state space were highlighted towards the enforcement of the agent policy. This is similar to a heat map visualization like Dombrowski et al. use except that only the input layer weights are considered \cite{dombrowski2019explanations}. The single hidden layer is a composition of the input features and is hence omitted from the interpretability analysis. We take the example of the top 2 largest weights for lockshop and the percentage of test cases that this held true in, to illustrate that the interpretability. The top 2 features are feature IDs 1 and 3 with 0.85\% test cases passed. We do this for the combined knowledge infusion setting. The feature IDs are from Table \ref{tab:relational-features}. We can see that for example:

\noindent for lockshop,  the features with ID $1$ and $3$,
hold the most weight for $\mathbf{85}$\% of the test cases.\\
\indent The simulator dynamics encodes that persons in a residential area shop at a “shop” that is in the neighborhood. This is what the “same” predicate means. Thus the feature states that there are many people in homes that are part of a residential area with a shop. Thus this implies due to the simulator dynamics that many people will be shopping regularly at this shop (i.e. high interaction). Therefore this result is not only interpretable but shows that the agent learns to implement locking down of a shop only when the shop is a source of high interaction among people and when it is open. 
\section{Conclusion and Future Work}
In settings where there is continuously evolving dynamics and the resulting non-stationarity and partial observability, standard RL frameworks suffer from unaffordable delays due to non-stationarity and sub-optimal policy learning due to partial observability. This is because of their fundamental trial and error based correction. In our example setting of APC, this delay and in correction or a sub-optimal policy can prove extremely costly. In various other Real-Life Scenarios, we see similar issues. We develop a principled Knowledge Infusion framework to enable effective control of unintended consequences that arise there-of and demonstrate its effectiveness. We will explore more specifications for the knowledge as functional constraints and their applications in future work in enhanced simulation settings. 

\bibliography{ref}

\end{document}